\documentclass[letterpaper]{article} 
\usepackage[preprint]{aaai2027}
\usepackage[hyphens]{url}  
\usepackage{graphicx} 
\usepackage{natbib}  
\usepackage{caption} 
\usepackage{algorithm}
\usepackage{algorithmic}
\usepackage{amsmath}
\usepackage{amssymb}
\usepackage{multirow}
\usepackage{newfloat}
\usepackage{listings}
\DeclareCaptionStyle{ruled}{labelfont=normalfont,labelsep=colon,strut=off} 
\floatstyle{ruled}
\newfloat{listing}{tb}{lst}{}
\floatname{listing}{Listing}

\usepackage{booktabs}

\title{SpikeCredit: Temporal Credit Carrier for Reinforcement Learning with Sparse Rewards}

\author{
    Yingchao Yu\textsuperscript{\rm 1,2},
    Pengfei Sun\textsuperscript{\rm 3},
    Wenxuan Pan\textsuperscript{\rm 2},
    Wei Chen\textsuperscript{\rm 2},
    Yitian Hong\textsuperscript{\rm 4},
    Kuangrong Hao\textsuperscript{\rm 1},
    Yaochu Jin\textsuperscript{\rm 2}\corresponding
}

\affiliations{
    \textsuperscript{\rm 1}
    School of Information and Intelligent Science,
    Donghua University, Shanghai, China\\
    \textsuperscript{\rm 2}
    School of Engineering,
    Westlake University, Hangzhou, China\\
    \textsuperscript{\rm 3}
    Department of Electrical and Electronic Engineering,
    Imperial College London, London, United Kingdom\\
    \textsuperscript{\rm 4}
    School of Information Science and Engineering,
    East China University of Science and Technology, Shanghai, China\\
    yingchaoyu@mail.dhu.edu.cn,
    p.sun@imperial.ac.uk,
    panwenxuan@westlake.edu.cn,
    chenwei06@westlake.edu.cn,
    ythong1314@mail.ecust.edu.cn,
    krhao@dhu.edu.cn,
    jinyaochu@westlake.edu.cn
}

\begin{document}
\maketitle

\begin{abstract}
Reinforcement learning (RL) with sparse rewards is challenging because delayed outcomes provide little guidance about which intermediate computations caused success or failure. We argue that reliable credit assignment requires policy dynamics that preserve and expose credit-relevant information over time, a role we formalize as \textbf{Temporal Credit Carriers (TCCs)} and that spiking neural networks (SNNs) naturally fulfill through graded membrane traces and event-driven spikes. Based on this hypothesis, we propose \textbf{SpikeCredit}, an SNN-based framework for RL with sparse rewards that first performs task-adaptive TCC selection and then closes the loop between a fast TCC-reading pathway, where self-motion feedback constraint uses local behavior-grounded cues to constrain transition-level credit recovery, and a slow TCC-writing pathway, where credit-targeted trace alignment feeds recovered credit back into the actor to make future TCC dynamics more credit-readable. Across sparse-reward MuJoCo tasks, SpikeCredit improves Last10 return over sparse SNN baselines by +1169\% on Ant, +953\% on Hopper, +723\% on Swimmer, and +1781\% on Walker2d, and exceeds the dense-reward baseline on Swimmer by +113\%. Mechanistic analyses further show substantially stronger alignment with dense rewards than the sparse SNN baseline. These results position spiking dynamics as credit-preserving substrates for sparse-reward RL.
\end{abstract}


\section{Introduction}
Sparse rewards remain a fundamental challenge in reinforcement learning (RL), particularly in long-horizon control \cite{pignatelli2024survey,sutton1998reinforcement}. When meaningful feedback is delayed until the end of a trajectory, the learner knows whether the overall behavior succeeded or failed but receives little guidance about which intermediate states, actions, or computations were responsible. This creates a temporal credit-assignment problem: many different step-wise explanations are compatible with the same episodic outcome, making it difficult to distinguish causally useful behavior from incidental correlations. The resulting weak learning signal impedes exploration and forces value estimates or policy gradients to propagate information across long temporal distances \cite{andrychowicz2017hindsight,arjona2019rudder}.

\begin{figure}[t!]
\centering
\includegraphics[width=0.98\columnwidth]{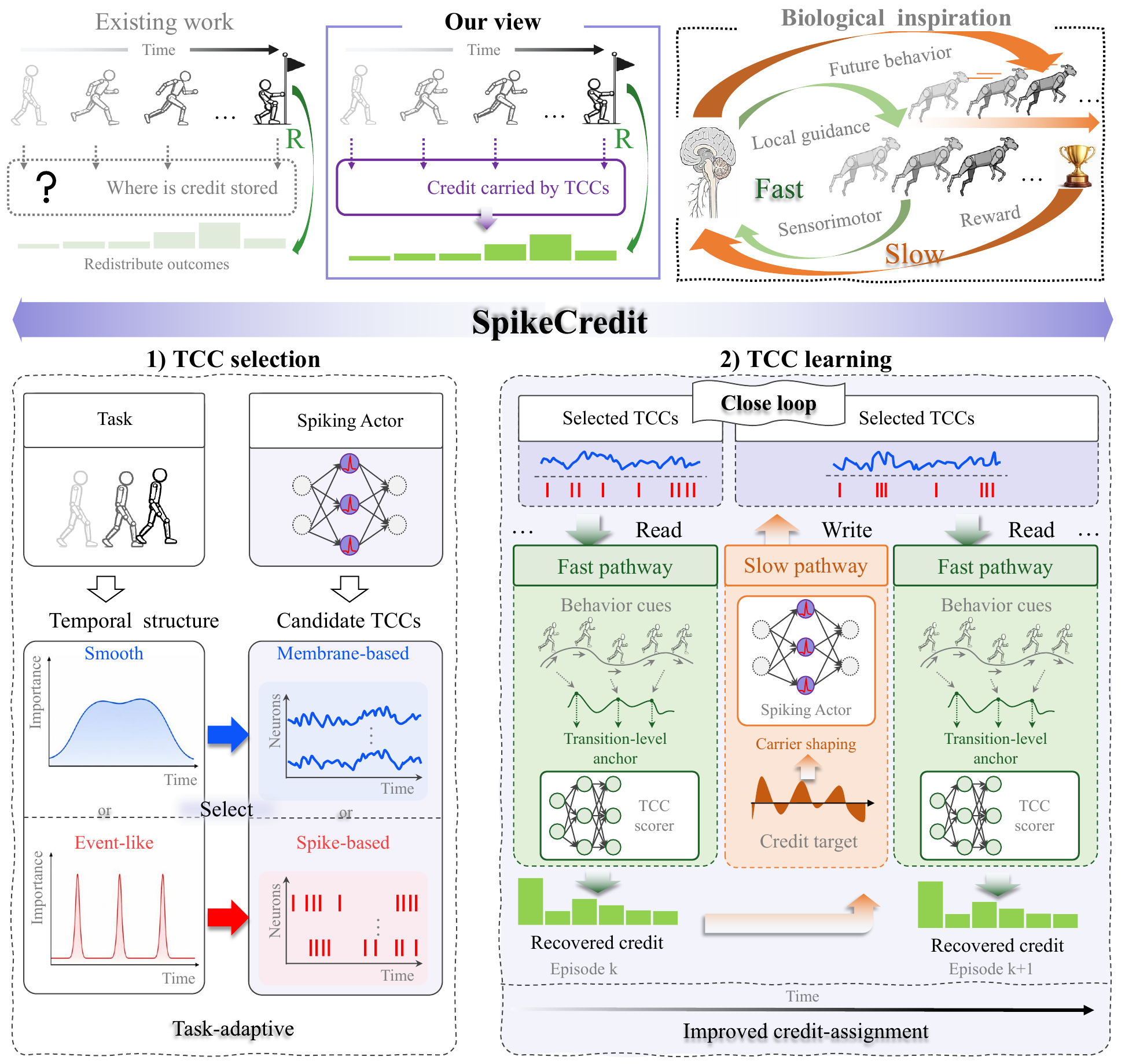} 
\caption{Motivation and core idea of SpikeCredit. Inspired by biological fast-slow learning, SpikeCredit treats spiking actor dynamics as task-adaptive temporal credit carriers (TCCs): the fast pathway reads credit from current dynamics, while the slow pathway writes recovered targets to reshape future dynamics, forming a closed read-write loop.}
\label{fig1}
\end{figure}

A broad line of work addresses this difficulty by constructing more informative supervision from sparse outcomes. Hindsight methods relabel past experience to expose otherwise unavailable successes \cite{andrychowicz2017hindsight}; return-decomposition and reward-redistribution methods assign delayed outcomes to selected portions of a trajectory \cite{arjona2019rudder,patil2022align,kapoor2025tar2}; and reward-shaping approaches learn auxiliary reward functions or models that provide denser feedback \cite{ng1999policy,ma2024assistant,venugopal2026occupancy}. Despite their different implementations, these approaches primarily focus on how to infer, redistribute, or reconstruct transition-level learning signals from trajectory-level outcomes. However, a delayed outcome contains little temporal structure by itself and cannot recover transition-level information that was not preserved in the policy or its interaction history. Rather than asking only how delayed outcomes should be redistributed, we ask a prior question: \emph{where is the credit-relevant temporal information required for reliable redistribution preserved in the first place?}

We posit that reliable credit recovery depends on policy dynamics that preserve and expose credit-relevant structure. We define such internal dynamics as \textbf{Temporal Credit Carriers (TCCs)}: neural states whose temporal evolution preserves information that can later support transition-level credit inference from delayed outcomes. Spiking neural network (SNN) actors provide a particularly suitable instantiation of TCCs through two complementary temporal dynamics. Membrane potentials integrate information over time, providing graded temporal traces that can preserve slowly evolving behavioral dependencies relevant to credit assignment. By contrast, spikes provide sparse and localized markers of salient behavioral events, which may expose event-specific temporal evidence for credit assignment.

Building on this insight, we propose \textbf{SpikeCredit}, a credit-aware temporal dynamics learning framework for sparse-reward RL, as illustrated in Figure~\ref{fig1}. SpikeCredit instantiates TCCs through task-adaptive selection between membrane traces and spike events according to the temporal characteristics of the task. However, selecting a suitable carrier is insufficient: a useful TCC should not only preserve temporal information, but also make such information readable for delayed credit inference and become adaptable through learning. SpikeCredit therefore couples two complementary operations: reading transition-level credit from current TCC dynamics and writing the recovered signals back to shape future dynamics.

This read-write mechanism draws inspiration from the complementary timescales of biological motor learning: rapid sensorimotor feedback constrains ongoing movement and provides local behavioral guidance, whereas delayed reward-related teaching signals reshape future control \cite{wolpert2011principles,scott2004optimal,schultz1997neural}. We translate this principle into a closed loop between credit reading and credit writing. The fast pathway, implemented by \textbf{Self-Motion Feedback Constraint (SMF)}, uses local behavior-grounded cues as temporal anchors to constrain credit inference from the actor's current TCC dynamics. The slow pathway, implemented by \textbf{Credit-Targeted Trace Alignment (CTT)}, feeds recovered credit targets back into the actor, shaping future TCC dynamics to become more readable for credit assignment. The updated dynamics are then reused by SMF in subsequent episodes, forming a closed loop in which credit inference and representation shaping mutually reinforce each other. This read-write loop reduces the ambiguity of return-only credit assignment while progressively improving future temporal credit recovery.
To sum up, our contributions are as follows: 
\begin{itemize}
    \item We formulate \textbf{Temporal Credit Carriers (TCCs)}, which consider internal policy dynamics as learnable substrates for preserving and exposing temporal evidence required for credit assignment. We instantiate TCCs in SNN actors through complementary membrane traces and spike events.
    \item We propose \textbf{SpikeCredit}, a fast-read and slow-write framework in which Self-Motion Feedback Constraint (SMF) uses behavior-grounded temporal anchors to recover credit from current TCC dynamics, while Credit-Targeted Trace Alignment (CTT) feeds the recovered targets back into the actor to improve the credit readability of future dynamics.
    \item Experiments on four sparse-reward MuJoCo tasks show consistent gains over sparse-reward artificial neural network (ANN) and SNN baselines. Mechanistic analyses and controlled ablations further demonstrate meaningful alignment with dense-reward structure and verify the complementary roles of carrier selection, SMF, and CTT.
\end{itemize}

\begin{figure*}[t]
\centering
\includegraphics[width=0.98\textwidth]{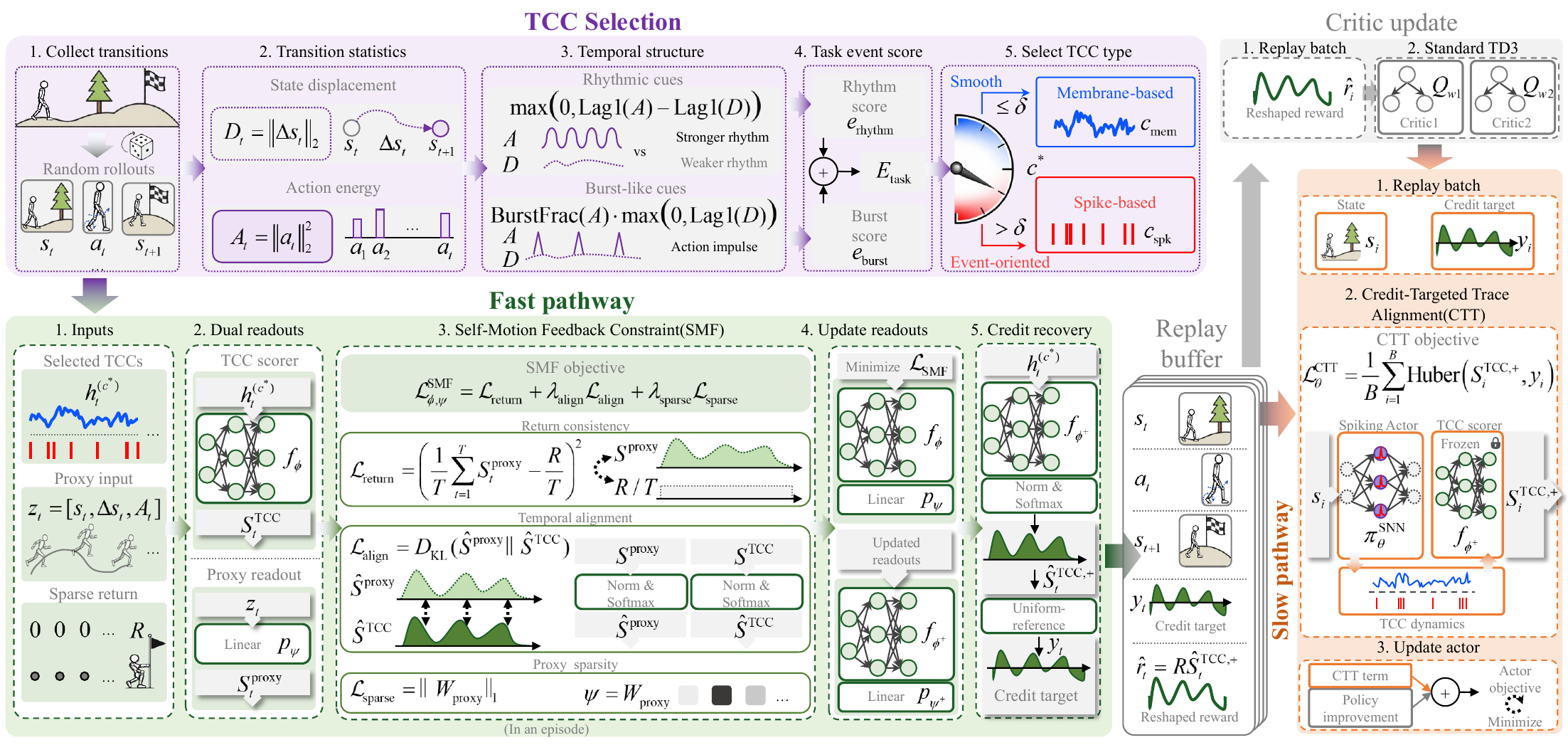} 
\caption{Overview of SpikeCredit. Task statistics select membrane- or spike-based TCCs. The fast pathway uses Self-Motion Feedback Constraint (SMF) to read credit from the selected dynamics and generate reshaped rewards and replay targets, while the slow pathway uses Credit-Targeted Trace Alignment (CTT) to write these targets back into the actor, progressively improving future TCC readability.}
\label{fig2}
\end{figure*}

\section{Related Work}
\subsection{Sparse-Reward RL}
Sparse-reward RL methods intervene at different stages of the learning process: exploration methods improve the discovery of rewarding trajectories \cite{pathak2017curiosity,burda2019rnd,pathak2019disagreement,sekar2020plan2explore}; demonstration, curriculum, and subgoal approaches reduce the difficulty of reaching them \cite{salimans2018montezuma,wilcox2022mcac,florensa2018goalgan,chen2021vacl}; hindsight relabeling reinterprets failed experience \cite{andrychowicz2017hindsight,fang2019curriculum,li2020generalized,pitis2020mega}; and reward shaping provides denser optimization signals \cite{ng1999policy,devidze2022exploration,ma2024assistant,ma2025dual,venugopal2026occupancy}. Temporal credit-assignment and reward-redistribution methods further identify which earlier decisions contributed to delayed outcomes \cite{arjona2019rudder,harutyunyan2019hindsight,patil2022align,han2022delayed,kapoor2025tar2}. 
By contrast, SpikeCredit focuses on whether internal policy dynamics contain credit-relevant information. It selects a task-adaptive TCC and refines its dynamics through a closed credit read-write loop, rather than relying on exploration bonuses or standalone reward redistribution.

\subsection{Spiking RL}
Early spiking RL focuses on how delayed rewards or temporal-difference signals modulate local synaptic plasticity through eligibility traces and reward-modulated spike-timing-dependent plasticity \cite{izhikevich2007distal,florian2007reinforcement,potjans2009actorcritic,fremaux2013continuous}. Deep spiking RL then targets scalable continuous control, using population coding, surrogate-gradient optimization, dynamic neurons, and fully spiking action readouts to improve the representation and deployment of spiking policies \cite{tang2021population,tan2021strategy,zhang2022multiscale,chen2025fully}. More recent work tailors target updates, gradient estimation, recurrent memory, and normalization to the discrete temporal dynamics and non-stationary training conditions of spiking RL \cite{xu2025proxy,vandenberghe2025adaptive,qin2025gated,xu2026care}. Overall, these studies primarily address how spiking policies are learned, represented, stabilized, and deployed. 
By contrast, to the best of our knowledge, SpikeCredit is the first spiking RL framework to exploit the distinct temporal characteristics of membrane potentials and spike events for sparse-reward credit
assignment.

\section{Preliminaries}
\subsection{Sparse-Reward RL}
\label{subsec:sparse_rl}
A RL problem is typically modeled as a Markov Decision Process (MDP)
\(\mathcal{M}=(\mathcal{S},\mathcal{A},P,r,\gamma)\), where \(\mathcal{S}\), \(\mathcal{A}\), \(P\), \(r\), and \(\gamma\) denote the state space, action space, transition dynamics, reward function, and discount factor, respectively. A policy
\(\pi_\theta(a_t\mid s_t)\), together with \(P\), induces trajectories
\(\tau=(s_1,a_1,\ldots,s_T,a_T,s_{T+1})\) of \(T\) transitions.

We consider a sparse-reward setting in which the agent receives no intermediate rewards and only observes  a terminal trajectory-level outcome \(R\):
\begin{equation}
r_t =
\begin{cases}
0, & t<T,\\
R, & t=T.
\end{cases}
\label{eq:sparse_reward}
\end{equation}
This terminal-only supervision creates temporal credit ambiguity, since the same terminal outcome can be explained by multiple transition-level credit assignments.

\subsection{Spiking Actor}
\label{subsec:snn_actor}

We adopt an actor-critic framework whose actor is implemented by an SNN policy \(\pi_{\theta}^{\mathrm{SNN}}\) that maps a continuous state \(s_t\) to an action \(a_t\):
\begin{equation}
a_t
=
\pi_{\theta}^{\mathrm{SNN}}(s_t).
\label{eq:snn_policy}
\end{equation}

The spiking actor consists of a spiking encoder, a spiking MLP, and a spiking decoder. Given \(s_t\), the encoder first converts it into a population-coded spike train:
\begin{equation}
x_t
=
\mathrm{Enc}^{\mathrm{SNN}}(s_t),
\label{eq:snn_encoding}
\end{equation}
where \(x_t\in\{0,1\}^{d_x\times K}\), \(d_x\) is the encoding dimension, and \(K\) is the number of SNN simulation steps within each environment transition.
The encoded spike train is then processed by the spiking MLP:
\begin{equation}
o_t
=
\mathrm{MLP}^{\mathrm{SNN}}(x_t),
\label{eq:snn_mlp}
\end{equation}
where \(o_t\) denotes its output spike train. 
In addition to producing \(o_t\), the hidden layers of the spiking MLP
expose temporally structured internal dynamics, including membrane-potential traces \(m_t\in\mathbb{R}^{d_h\times K}\) and spike events
\(e_t\in\{0,1\}^{d_h\times K}\), where \(d_h\) denotes the hidden feature dimension.
We treat these dynamics as candidate TCCs. Specifically, a TCC state is obtained by selecting a carrier type
\(c\in\mathcal{C}=\{c_{\mathrm{mem}},c_{\mathrm{spk}}\}\):
\begin{equation}
h_t^{(c)}
=
\begin{cases}
m_t, & c=c_{\mathrm{mem}},\\
e_t, & c=c_{\mathrm{spk}},
\end{cases}
\label{eq:candidate_tcc}
\end{equation}
where \(c_{\mathrm{mem}}\) and \(c_{\mathrm{spk}}\) denote the membrane-based and spike-based carrier types, respectively.
Finally, the decoder converts the output spike train into a continuous control action:
\begin{equation}
a_t
=
a_{\max}
\odot
\mathrm{Dec}^{\mathrm{SNN}}(o_t),
\label{eq:snn_action}
\end{equation}
where \(a_{\max}\) denotes the action bound and \(\odot\) denotes element-wise multiplication.

\section{Methodology}
As shown in Figure~\ref{fig2}, SpikeCredit addresses sparse-reward credit assignment by treating spiking actor dynamics as task-adaptive Temporal Credit Carriers (TCCs). It first selects a suitable carrier, then uses Self-Motion Feedback Constraint (SMF) to recover transition-level credit and Credit-Targeted Trace Alignment (CTT) to write the recovered targets back into the actor, forming a closed read-write loop that progressively improves the credit readability of future TCC dynamics.

\subsection{Task-Adaptive TCC Selection}
\label{subsec:carrier_selection}
SpikeCredit selects a suitable TCC type according to the temporal structure of each task. We consider two complementary carrier types:
\begin{equation}
\mathcal{C}
=
\{c_{\mathrm{mem}},c_{\mathrm{spk}}\},
\label{eq:carrier_set}
\end{equation}
where membrane traces provide graded temporal integration, whereas spike events provide localized event markers. Different temporal structures may therefore favor different carrier types.

During TCC selection, we collect a small number of random rollouts and record the transition tuples \((s_t,a_t,s_{t+1})\). For each transition, we compute
\begin{equation}
D_t
=
\|\Delta s_t\|_2,
\qquad
A_t
=
\|a_t\|^2_2,
\label{eq:transition_signals}
\end{equation}
where \(\Delta s_t=s_{t+1}-s_t\). Here, \(D_t\) measures local state displacement and \(A_t\) measures action energy.

We then define the task event score as
\begin{equation}
E_{\mathrm{task}}
=
e_{\mathrm{rhythm}}
+
e_{\mathrm{burst}},
\label{eq:raw_event_score}
\end{equation}
where
\begin{equation}
e_{\mathrm{rhythm}}
=
\max\left(
0,\,
\mathrm{Lag1}(A)-\mathrm{Lag1}(D)
\right),
\label{eq:rhythm_positive}
\end{equation}
\begin{equation}
e_{\mathrm{burst}}
=
\mathrm{BurstFrac}(A)
\cdot
\max\left(
0,\,
\mathrm{Lag1}(D)
\right).
\label{eq:burst_score}
\end{equation}
The rhythm component captures periodic structures in action dynamics, while the burst component captures localized action events under coherent state transitions. \(\mathrm{BurstFrac}(x)\) denotes the fraction of local peaks above threshold in the normalized sequence.

The carrier type is selected by:
\begin{equation}
c^*
=
\begin{cases}
c_{\mathrm{spk}}, & E_{\mathrm{task}}>\delta,\\
c_{\mathrm{mem}}, & E_{\mathrm{task}}\leq\delta,
\end{cases}
\label{eq:selected_carrier}
\end{equation}
where \(\delta\) is a fixed selection threshold.
A larger \(E_{\mathrm{task}}\) indicates stronger rhythmic or event-oriented temporal structure and favors spike-based carriers. Otherwise, membrane-based carriers are selected for smoother temporal dynamics. This selection stage is training-free.

\subsection{Fast Pathway: SMF for TCC Reading}
\label{subsec:smf}
After selecting the task-adaptive TCC type \(c^*\), SpikeCredit collects an episode and records the corresponding selected TCC dynamics \(\{h_t^{(c^*)}\}_{t=1}^{T}\). 
To estimate transition-level credit from these dynamics, we introduce a lightweight TCC scorer \(f_{\phi}\):
\begin{equation}
S_t^{\mathrm{TCC}}
=
f_{\phi}\!\left(h_t^{(c^*)}\right).
\label{eq:smf_tcc_scorer}
\end{equation}
However, the terminal outcome \(R\) alone does not provide sufficient information to uniquely determine transition-level credit. SMF therefore introduces behavior-grounded temporal anchors.
Specifically, we construct the self-motion proxy input
\begin{equation}
z_t=[s_t,\Delta s_t,A_t],
\end{equation}
A linear proxy readout produces a behavior-grounded transition score:
\begin{equation}
S_t^{\mathrm{proxy}}
=
p_{\psi}(z_t)
=
W_{\mathrm{proxy}}^{\top}z_t,
\qquad
\psi=W_{\mathrm{proxy}}.
\label{eq:smf_proxy_out}
\end{equation}
This proxy does not provide an additional reward signal. Instead, it acts as a temporal constraint that guides credit inference from the TCC dynamics.

SMF optimizes the temporal scores using
\begin{equation}
\mathcal{L}_{\mathrm{SMF}}
=
\mathcal{L}_{\mathrm{return}}
+
\lambda_{\mathrm{align}}\mathcal{L}_{\mathrm{align}}
+
\lambda_{\mathrm{sparse}}\mathcal{L}_{\mathrm{sparse}},
\label{eq:smf_loss}
\end{equation}
where \(\lambda_{\mathrm{align}}\) and \(\lambda_{\mathrm{sparse}}\) are loss coefficients.

The return-consistency term is
\begin{equation}
\mathcal{L}_{\mathrm{return}}
=
\left(
\frac{1}{T}\sum_{t=1}^{T}S_t^{\mathrm{proxy}}
-
\frac{R}{T}
\right)^2.
\label{eq:smf_return_loss}
\end{equation}
This term encourages the proxy scores to aggregate to the observed episodic outcome while keeping the loss scale less sensitive to the trajectory length.

Before temporal alignment, we transform the proxy and TCC score sequences into comparable trajectory-level distributions:
\begin{equation}
\hat S^{\mathrm{proxy}}
=
\mathrm{Softmax}
\left(
\mathrm{Norm}\left(\{S_t^{\mathrm{proxy}}\}_{t=1}^{T}\right)/\tau_f
\right),
\label{eq:smf_distribution_p}
\end{equation}
\begin{equation}
\hat S^{\mathrm{TCC}}
=
\mathrm{Softmax}
\left(
\mathrm{Norm}\left(\{S_t^{\mathrm{TCC}}\}_{t=1}^{T}\right)/\tau_f
\right),
\label{eq:smf_distribution_h}
\end{equation}
where \(\tau_f\) is a temperature parameter and \(\mathrm{Norm}(\cdot)\) denotes normalization over the trajectory.
The temporal alignment objective is
\begin{equation}
\mathcal{L}_{\mathrm{align}}
=
D_{\mathrm{KL}}
\left(
\hat S^{\mathrm{proxy}}
\middle\|
\hat S^{\mathrm{TCC}}
\right).
\label{eq:smf_align_loss}
\end{equation}

We additionally impose sparsity on the proxy readout:
\begin{equation}
\mathcal{L}_{\mathrm{sparse}}
=
\|W_{\mathrm{proxy}}\|_1.
\label{eq:smf_sparse_loss}
\end{equation}
This regularization encourages the proxy to rely on a compact subset of self-motion features rather than behaving as an unconstrained predictor. 
Together, \(\mathcal{L}_{\mathrm{return}}\) grounds the proxy scores in the episodic outcome,
\(\mathcal{L}_{\mathrm{align}}\) transfers their behavior-grounded temporal structure to the TCC scorer, and \(\mathcal{L}_{\mathrm{sparse}}\) keeps the proxy interpretable and
compact.

After each episode, we hold the actor fixed and update \((\psi,\phi)\) by minimizing \(\mathcal{L}_{\mathrm{SMF}}\). Using the updated scorer \(f_{\phi^+}\), we then recompute the TCC score sequence:
\begin{equation}
S^{\mathrm{TCC},+}
=
\left\{
f_{\phi^+}\!\left(h_t^{(c^*)}\right)
\right\}_{t=1}^{T}.
\label{eq:updated_tcc_score}
\end{equation}
A direction-aware trajectory-level credit distribution is obtained as
\begin{equation}
\hat S^{\mathrm{TCC},+}
=
\mathrm{Softmax}
\left(
\mathrm{sgn}(R)\,
\mathrm{Norm}\left(S^{\mathrm{TCC},+}\right)/\tau_s
\right),
\label{eq:ctt_credit_distribution}
\end{equation}
where \(\tau_s\) is a temperature parameter. For \(R>0\), higher-scoring transitions receive larger credit weights; for \(R<0\), the ordering is reversed, assigning larger weights to lower-scoring transitions.

The terminal outcome is redistributed as
\begin{equation}
\hat r_t
=
R\hat S_t^{\mathrm{TCC},+}.
\label{eq:redistributed_reward}
\end{equation}
We further construct a uniform-referenced log-credit target:
\begin{equation}
y_t
=
\log\left(T\hat S_t^{\mathrm{TCC},+}\right).
\label{eq:ctt_credit_target}
\end{equation}
Since \(1/T\) is the uniform credit level, \(y_t>0\), \(y_t=0\), and \(y_t<0\) indicate above-uniform, uniform, and below-uniform credit, respectively. Thus, \(y_t\) specifies whether the corresponding transition should produce a stronger or weaker credit-readable TCC than the uniform trajectory reference. The redistributed reward \(\hat r_t\) and credit target \(y_t\) are stored with the corresponding transition in the replay buffer.

\subsection{Slow Pathway: CTT for TCC Writing}
\label{subsec:ctt}
Credit-Targeted Trace Alignment (CTT) uses SMF-recovered credit targets to reshape future TCC dynamics for credit assignment.

During actor updates, replayed states \(s_i\) are passed through the current spiking actor to generate new selected TCC dynamics \(h_{\theta}^{(c^*)}(s_i)\). A frozen TCC scorer \(f_{\phi^+}\), where \(\phi^+\) denotes the scorer parameters updated by the fast pathway and held fixed during the actor update, maps these dynamics to current TCC scores. We align these scores with the stored credit targets \(y_i\) using the Huber loss \cite{huber1964robust}:
\begin{equation}
\mathcal{L}^{\mathrm{CTT}}_{\theta}
=
\frac{1}{B}
\sum_{i=1}^{B}
\mathrm{Huber}
\left(
f_{\phi^+}
\left(
h_{\theta}^{(c^*)}(s_i)
\right),
y_i
\right),
\label{eq:ctt_loss}
\end{equation}
where \(B\) is the replay batch size. Freezing the scorer ensures that the recovered credit remains a fixed shaping target, so gradients from \(\mathcal{L}_{\mathrm{CTT}}\) update only the actor rather than altering the credit estimator.

Through this slow replay-based update, credit targets inferred after each episode are written back into the actor's internal representation. Consequently, future TCC dynamics become more informative for the frozen scorer to read, closing the credit loop from post-hoc TCC reading to credit-guided TCC writing.

\subsection{Closed-Loop Optimization}
\label{subsec:closed_loop_optimization}

\begin{algorithm}[t!]
\caption{Closed-Loop Optimization in SpikeCredit}
\label{alg:spikecredit}

\begin{algorithmic}[1]
\STATE Select the task-adaptive TCC type \(c^*\) and initialize replay buffer \(\mathcal{D}\).
\FOR{each episode}
    \STATE Collect a trajectory
    \(\mathcal{B}_{\tau}
    =
    \{(s_t,a_t,s_{t+1},d_t,z_t,h_t^{(c^*)})\}_{t=1}^{T}\),
    where \(z_t=[s_t,\Delta s_t,A_t]\).
    \STATE Observe the terminal outcome \(R\).

    \STATE \textbf{Fast pathway: read credit from current TCCs}
    \STATE Hold the actor fixed and update \((\psi,\phi)\) by minimizing
    \(\mathcal{L}_{\mathrm{SMF}}\) on \(\mathcal{B}_{\tau}\).
    \STATE Use \(f_{\phi^+}\) to obtain
    \(\hat S^{\mathrm{TCC},+}\), and compute
    \[
    \hat r_t=R\hat S_t^{\mathrm{TCC},+},
    \qquad
    y_t=\log(T\hat S_t^{\mathrm{TCC},+}).
    \]
    \STATE Store
    \((s_t,a_t,s_{t+1},\hat r_t,y_t,d_t)\)
    in \(\mathcal{D}\).

    \STATE \textbf{Slow pathway: write credit into future TCCs}
    \STATE Sample replay mini-batches from \(\mathcal{D}\) and update the twin critics using \(\hat r_i\).
    \STATE On delayed actor-update steps, freeze \(f_{\phi^+}\) and update the actor using
    \(\mathcal{L}_{\mathrm{actor}}\).
\ENDFOR
\end{algorithmic}
\end{algorithm}

Algorithm~\ref{alg:spikecredit} summarizes the closed fast-slow optimization loop. After each completed episode, the fast pathway reads a temporal credit distribution from the current TCC dynamics and produces redistributed rewards \(\hat r_t\) and credit targets \(y_t\). The slow pathway then uses \(\hat r_t\) to update the critics and aligns future TCC dynamics with \(y_t\) during delayed actor updates, progressively improving their credit readability.

\begin{figure}[t!]
\centering
\includegraphics[width=0.9\columnwidth]{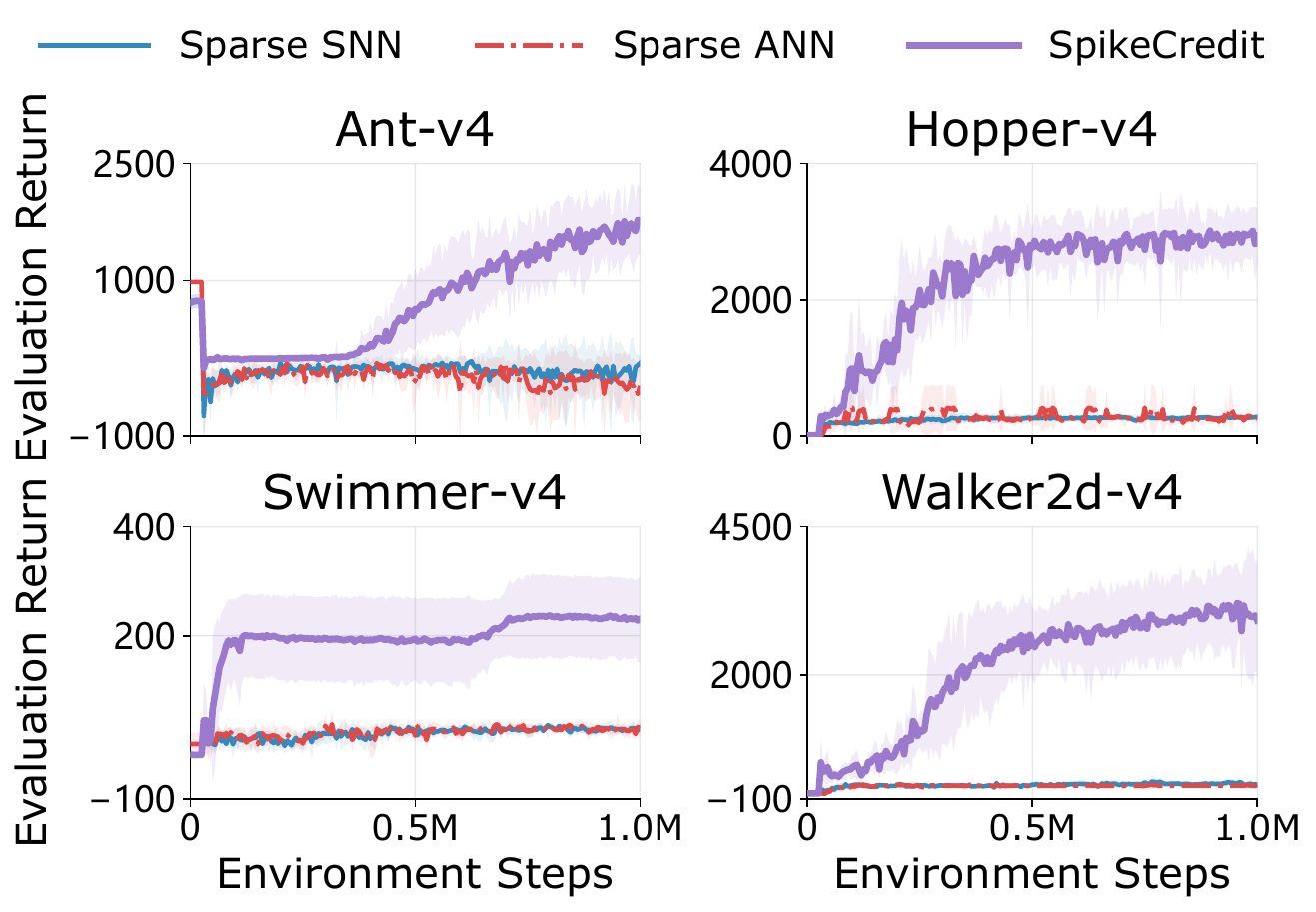} 
\caption{Learning curves of SpikeCredit, sparse-reward SNN and ANN baselines on four MuJoCo tasks. Lines and shaded regions denote the mean and standard deviation over five seeds, respectively.}
\label{res1}
\end{figure}

\begin{figure*}[h]
\centering
\includegraphics[width=0.9\textwidth]{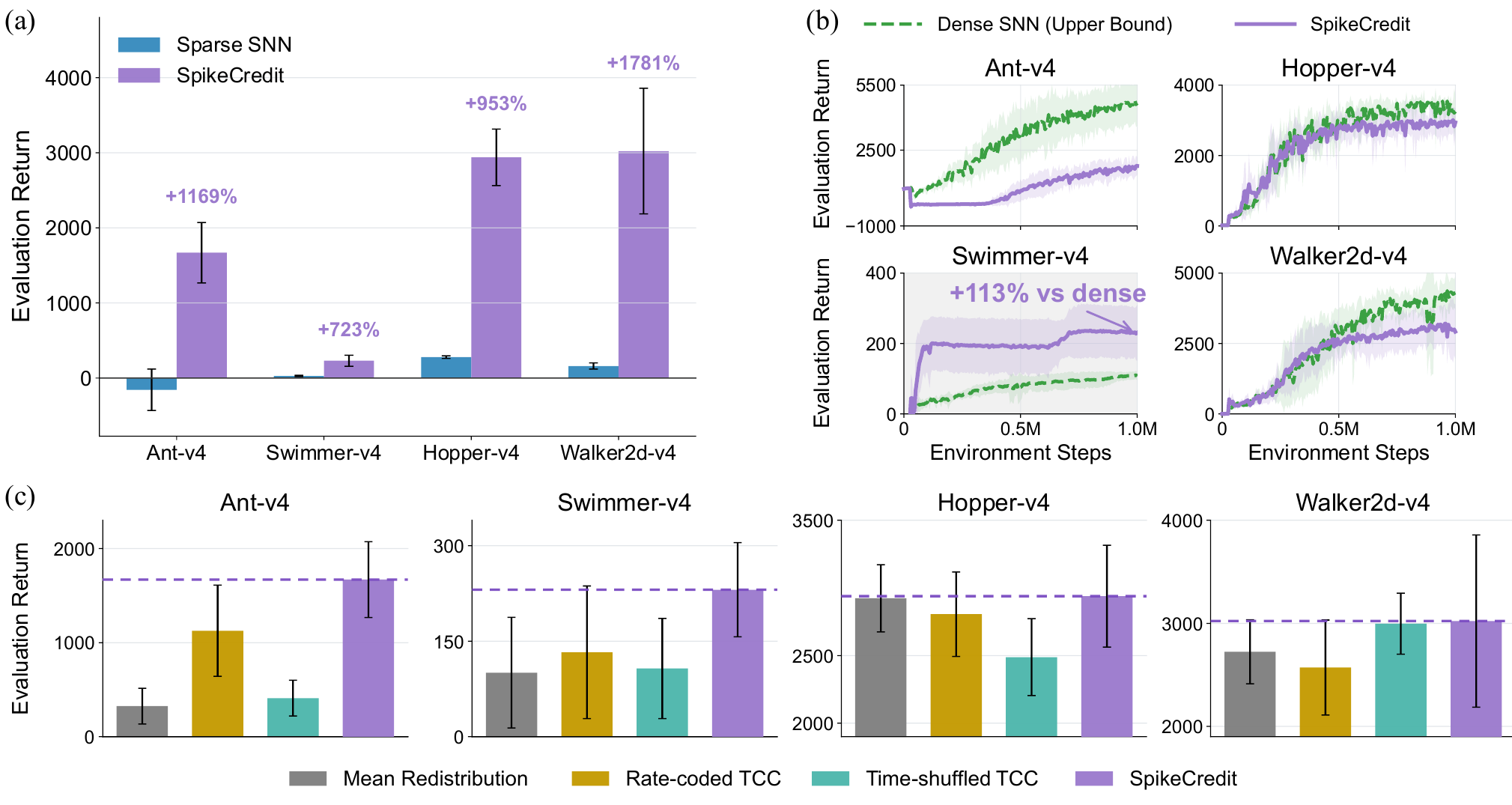} 
\caption{Overall performance on four MuJoCo tasks. (a) Last10 returns against the sparse SNN baseline. (b) Learning curves against the dense SNN upper bound. (c) Comparisons with three alternative credit-assignment methods. Mean \(\pm\) standard deviation over five seeds.}
\label{res1-2}
\end{figure*}

For a replay mini-batch \(\{(s_i,a_i,s_i',\hat r_i,d_i)\}_{i=1}^{B}\), we update the twin critics \(Q_{\omega_1}\) and \(Q_{\omega_2}\) following standard TD3 \cite{fujimoto2018addressing}, with the environment reward replaced by the redistributed reward \(\hat r_i\):
\begin{equation}
\mathcal{L}_{\mathrm{critic}}
=
\frac{1}{B}
\sum_{i=1}^{B}
\sum_{j=1}^{2}
\left(
Q_{\omega_j}(s_i,a_i)
-
q_i^{\mathrm{tar}}
\right)^2,
\label{eq:critic_loss}
\end{equation}
where \(q_i^{\mathrm{tar}}\) is the standard TD3 target computed using \(\hat r_i\).

On delayed policy-update steps, the actor objective combines standard policy improvement with CTT:
\begin{equation}
\mathcal{L}_{\mathrm{actor}}
=
-
\frac{1}{B}
\sum_{i=1}^{B}
Q_{\omega_1}
\left(
s_i,
\pi_{\theta}^{\mathrm{SNN}}(s_i)
\right)
+
\lambda_{\mathrm{CTT}}
\mathcal{L}^{\mathrm{CTT}}_{\theta},
\label{eq:ctt_actor_loss}
\end{equation}
where \(\lambda_{\mathrm{CTT}}\) controls the strength of TCC shaping.

\section{Experiment}

\subsection{Experimental Setup}
We evaluate SpikeCredit on four MuJoCo benchmarks \cite{todorov2012mujoco,todorov2014convex} from Gymnasium \cite{brockman2016openai,towers2024gymnasium}: Ant \cite{schulman2015high}, Hopper \cite{erez2012infinite}, Swimmer \cite{coulom2002reinforcement}, and Walker2d. These tasks cover diverse control characteristics, including quadrupedal coordination, periodic hopping, phase-coordinated propulsion, and bipedal locomotion. For the sparse setting, intermediate rewards are set to zero, and their undiscounted sum is revealed only at termination or truncation.
We adopt the recently proposed CaRe-BN \cite{xu2026care} as the backbone of SpikeCredit. ANN and SNN methods in this work share the same twin critics and differ only in the actor.
SMF is updated once per episode, whereas CTT is enabled only during delayed actor updates after warm-up. We report results over five random seeds.
Environment illustrations, network and spiking-neuron configurations, RL and SpikeCredit hyperparameters, and evaluation details are provided in the Appendix.

\subsection{Effective Credit Recovery with SpikeCredit}
We first evaluate SpikeCredit under trajectory-level sparse rewards. Figure~\ref{res1} shows sustained learning across all four tasks, while the sparse-reward SNN and ANN baselines remain at substantially lower returns. The Last10 results in Figure~\ref{res1-2}a confirm this advantage: relative to the sparse SNN, SpikeCredit improves the evaluation return by \(1169\%\), \(723\%\), \(953\%\), and \(1781\%\) on Ant, Swimmer, Hopper, and Walker2d, respectively. These gains show that the closed TCC read-write loop substantially improves sparse-reward learning beyond the underlying spiking actor. 
Compared with the same SNN trained on original dense rewards, Figure~\ref{res1-2}b shows that SpikeCredit approaches dense-reward performance on Hopper and Walker2d, despite receiving only one return per episode, while a larger gap remains on Ant. On Swimmer, it exceeds the dense-reward baseline by approximately \(113\%\), showing that sparse episodic supervision does not necessarily limit final performance when temporal credit is effectively recovered.

To verify that these gains arise from meaningful carrier structure rather than redistribution alone, Figure~\ref{res1-2}c compares SpikeCredit with three controlled alternatives. Uniform mean redistribution removes carrier information and transition-specific differentiation; rate-coded TCCs preserve aggregate spiking activity but discard fine-grained temporal dynamics; and time-shuffled TCCs preserve carrier values but break their trajectory alignment. SpikeCredit achieves the highest average performance on all four tasks, whereas no alternative performs consistently well across environments. Together, these results suggest that SpikeCredit's consistent advantage comes from jointly exploiting transition-specific credit information, fine-grained TCC dynamics, and their correct temporal alignment.

\begin{figure}[t!]
\centering
\includegraphics[width=0.95\columnwidth]{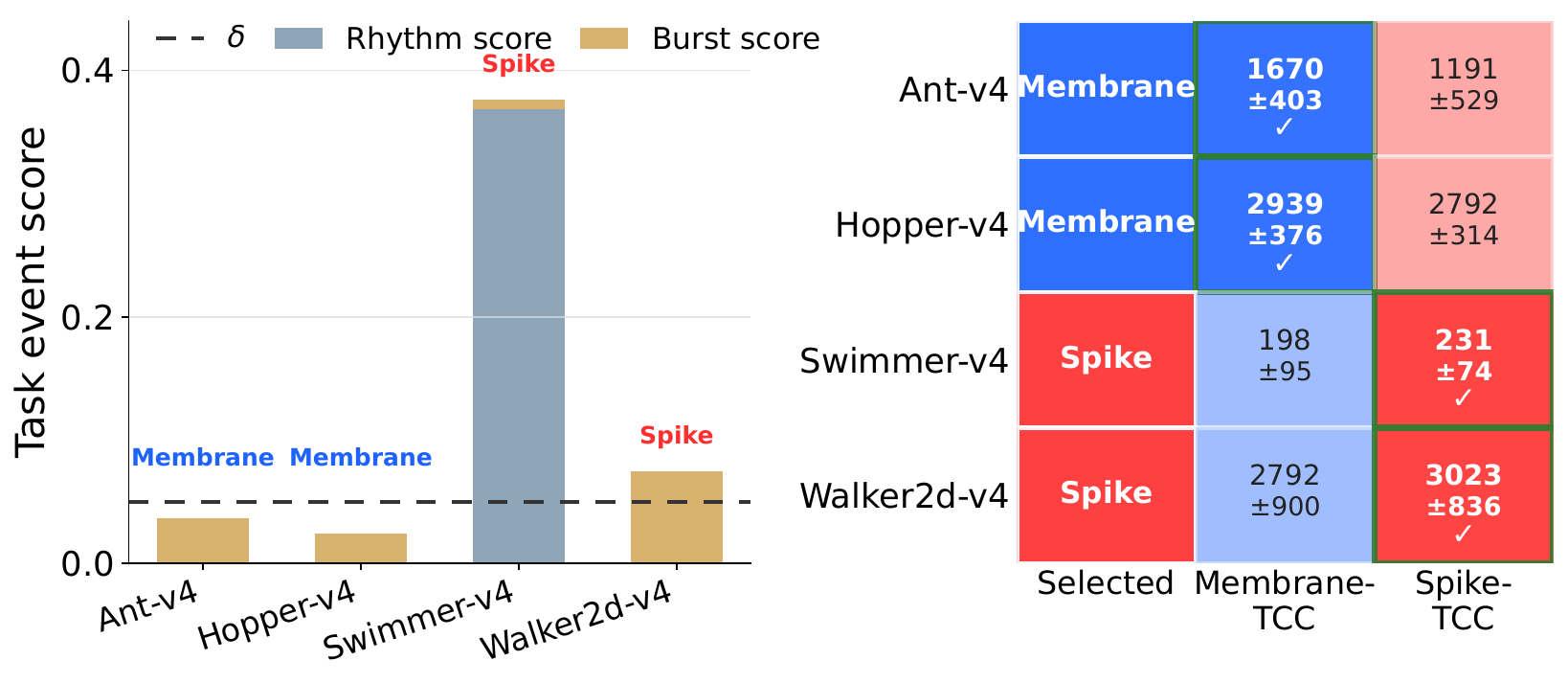} 
\caption{Task-adaptive TCC selection. Left: task event scores determine the carrier. Right: Last10 returns verify that the selected candidate performs best on each task. Results are averaged over five seeds.}
\label{res5}
\end{figure}

\begin{figure}[t!]
\centering
\includegraphics[width=0.92\columnwidth]{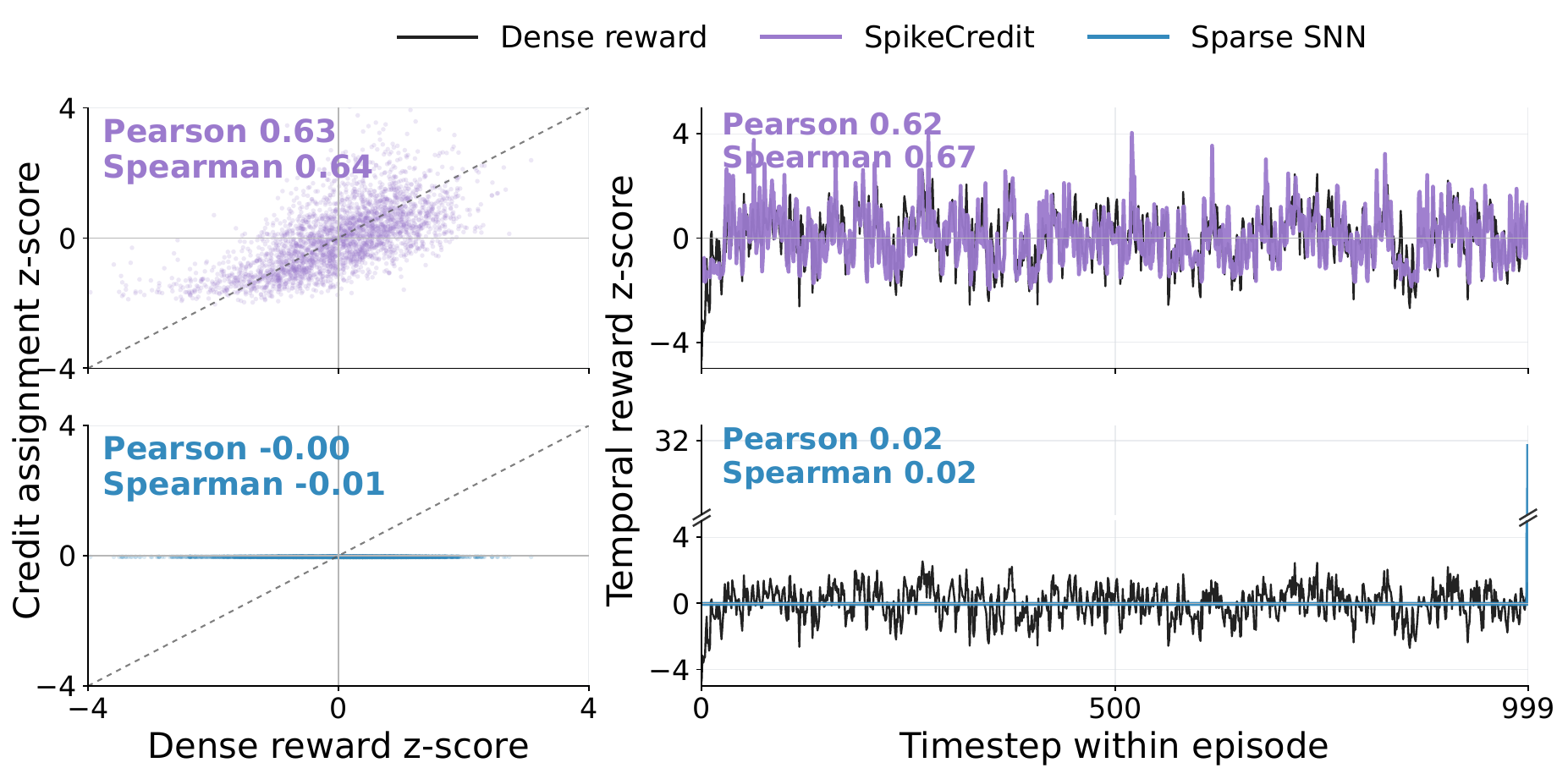} 
\caption{Alignment between credit signals and dense rewards on Ant-v4. Left: transition-wise correlations. Right: temporal profiles over an episode. Dense rewards are used only as a diagnostic reference.}
\label{res6}
\end{figure}

\subsection{Mechanistic Analysis}
\paragraph{Task-adaptive carrier selection.}
We first examine whether the task-event score can select a suitable credit carrier for each environment. As shown in Figure~\ref{res5}, Ant and Hopper exhibit event scores below the threshold and therefore select membrane traces, whereas Swimmer and Walker2d select spike events due to their stronger rhythmic or burst signatures. The selected carrier achieves the highest average return on all four tasks: membrane TCCs outperform spike TCCs on Ant and Hopper, while spike TCCs perform better on Swimmer and Walker2d. These results support the task-adaptive selection rule and show that no single carrier is uniformly optimal across different control dynamics. In particular, stronger event-level temporal structure favors spike-based carriers, whereas tasks with weaker event signatures benefit more from graded membrane dynamics.

\paragraph{Recovery of transition-level credit.}
We next examine whether SpikeCredit recovers meaningful temporal credit from episodic outcomes, using the original dense reward only as a diagnostic reference. As shown in Figure~\ref{res6}, its credit signals on Ant align well with dense rewards, with transition-wise Pearson/Spearman correlations of \(0.63/0.64\) and temporal-profile correlations of \(0.62/0.67\). By contrast, the sparse SNN remains nearly uncorrelated. These results show that SpikeCredit converts terminal-only feedback into temporally differentiated credit that tracks the dense-reward structure.

\begin{figure}[t!]
\centering
\includegraphics[width=0.92\columnwidth]{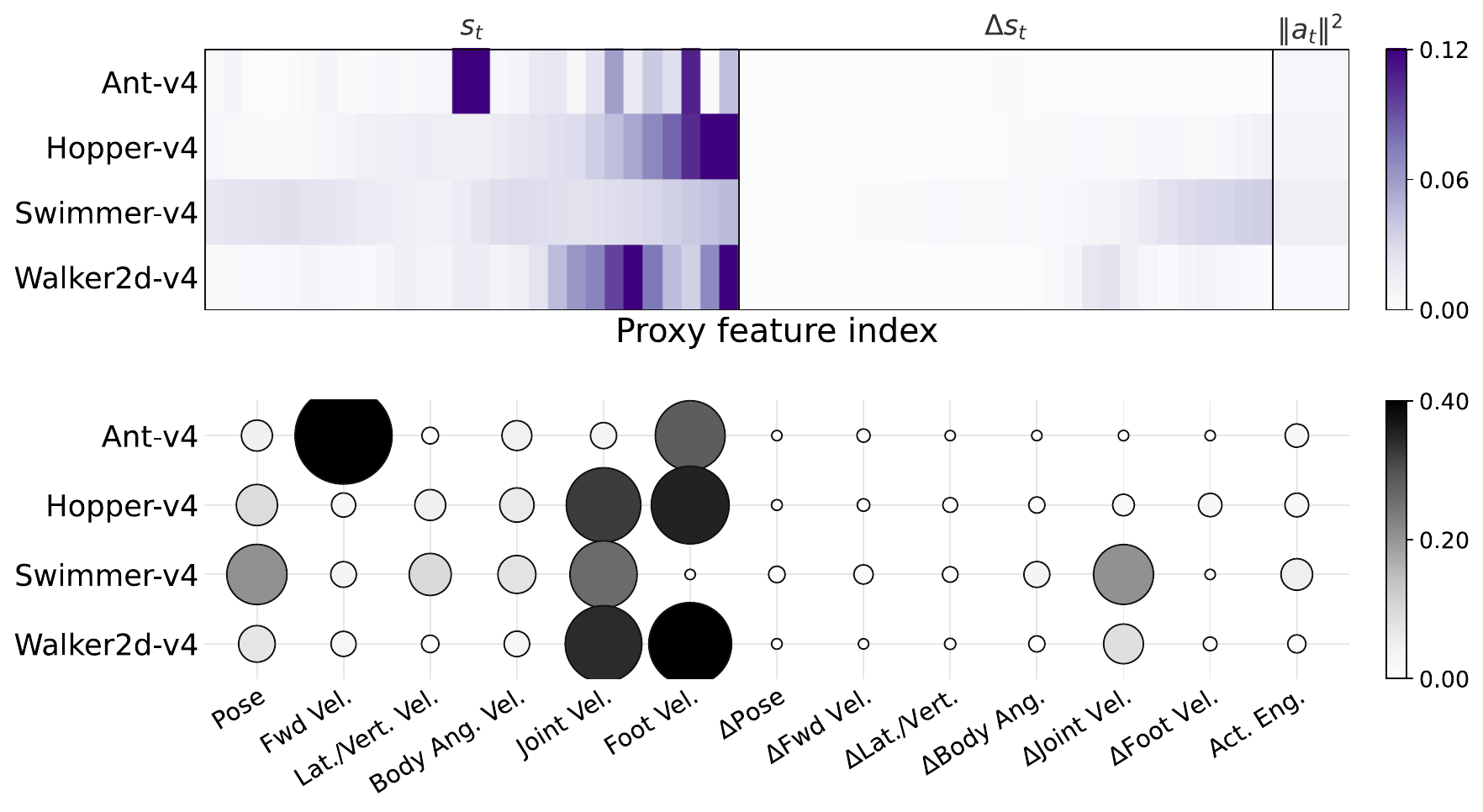} 
\caption{Task-dependent SMF proxy attribution. Top: feature importance over \(s_t\), \(\Delta s_t\), and \(\|a_t\|_2^2\). Bottom: attribution aggregated by physical feature groups. Semantic groupings of observation fields used for self-motion proxy attribution are provided in the Appendix.
}
\label{res8}
\end{figure}

\paragraph{Behavioral grounding of SMF.}
Figure~\ref{res8} shows that SMF relies mainly on state and state-change features, with limited contribution from action energy. The dominant cues are task-dependent: forward velocity on Ant, joint and foot velocities on Hopper and Walker2d, and pose and coordinated joint motion on Swimmer. Thus, SMF derives behavior-grounded temporal anchors from task-relevant self-motion cues rather than relying on a predefined motion rule.

\begin{table}[t!]
\centering
\begin{tabular}{@{}cccc@{}}
\toprule
Methods                         & Last10 Return     & Peak Return        & Peak Step \\ \midrule
SpikeCredit                     & $1669.9 \pm 402.8$ & $1777.5 \pm 466.8$ & 995K      \\
w/o $s_t$                 & $544.4 \pm 389.7$  & $738.1 \pm 91.5$   & 20K       \\
w/o $\Delta s_t$   & $1106.3 \pm 470.5$ & $1212.2 \pm 490.5$ & 985K      \\
w/o $\|a_t\|_2^2$ & $845.8 \pm 672.7$  & $939.5 \pm 654.9$  & 980K      \\
w/o SMF          & $192.7 \pm 78.6$   & $738.1 \pm 91.5$   & 20K       \\ \midrule
SpikeCredit                     & $1669.9 \pm 402.8$ & $1777.5 \pm 466.8$ & 995K      \\
w/o CTT                         & $1403.8 \pm 547.0$ & $1514.5 \pm 618.9$ & 975K      \\ \bottomrule
\end{tabular}
\caption{Ablation of SMF inputs and CTT on Ant-v4. Results are averaged over five seeds.}
\label{ablation}
\end{table}

\subsection{Ablation Analysis}

We ablate SMF and its behavioral cues \(z_t=[s_t,\Delta s_t,\|a_t\|_2^2]\) on Ant. As shown in Table~\ref{ablation}, SpikeCredit achieves a Last10 return of \(1669.9\pm402.8\). Removing \(s_t\), \(\Delta s_t\), or \(\|a_t\|_2^2\) reduces it to \(544.4\), \(1106.3\), and \(845.8\), respectively, while removing SMF causes it to collapse to \(192.7\pm78.6\). Moreover, the variants without \(s_t\) or SMF peak at only \(20\mathrm{K}\) steps, whereas the full model continues improving until nearly \(1\mathrm{M}\) steps. These results show that the three inputs provide complementary behavioral context, motion, and control-effort cues, with \(s_t\) being particularly important for sustained learning.
Removing CTT reduces the Last10 return from \(1669.9\) to \(1403.8\) and the peak return from \(1777.5\) to \(1514.5\), although learning continues until \(975\mathrm{K}\) steps. Overall, the ablations reveal complementary roles: SMF supports credit reading and sustained policy learning, while CTT further improves performance through credit writing.

\section{Conclusion}
We present SpikeCredit, a sparse-reward RL framework that uses spiking dynamics as temporal credit carriers within a closed read-write loop. Across four MuJoCo tasks, SpikeCredit consistently outperforms sparse-reward ANN and SNN baselines, while mechanistic analyses show that the recovered credit captures meaningful temporal structure.

\bibliography{aaai2027}
\clearpage
\twocolumn[
\begin{center}
    {\LARGE \textbf{Appendix}}
\end{center}
\vspace{1em}
]
\appendix

\section{Model Architecture}
\subsection{Dynamic neuron model}
\label{app:dynamic_neuron}

We use the second-order Dynamic Neuron (DN) model \cite{zhang2022multiscale} as the spiking unit in the SNN actor. Compared with the standard leaky integrate-and-fire (LIF) neuron \cite{burkitt2006review}, each DN maintains a membrane potential \(V\) and an adaptation variable \(U\), enabling richer internal dynamics during action generation.

For layer \(l\) and internal simulation step \(k\), the synaptic current is updated as
\begin{equation}
C_k^l = \alpha C_{k-1}^l + X_k^l,
\end{equation}
where \(X_k^l\) denotes the affine input to layer \(l\), and \(\alpha\) is the synaptic-current decay factor.

Before integrating the state at step \(k\), a spike emitted at the previous step resets the membrane potential and updates the adaptation variable:
\begin{equation}
\tilde V_{k-1}^l
=
(1-S_{k-1}^l)V_{k-1}^l
+
S_{k-1}^l\theta_r,
\end{equation}
\begin{equation}
\tilde U_{k-1}^l
=
U_{k-1}^l
+
S_{k-1}^l\theta_s,
\end{equation}
where \(\theta_r\) is the reset potential and \(\theta_s\) is the spike-triggered increment of \(U\).

The DN state is then updated using explicit Euler discretization with unit step size:
\begin{equation}
\Delta V_k^l
=
(\tilde V_{k-1}^l)^2
-
\tilde V_{k-1}^l
-
\tilde U_{k-1}^l
+
C_k^l,
\end{equation}
\begin{equation}
\Delta U_k^l
=
\theta_v \tilde V_{k-1}^l
+
\theta_u \tilde U_{k-1}^l,
\end{equation}
\begin{equation}
V_k^l
=
\tilde V_{k-1}^l+\Delta V_k^l,
\qquad
U_k^l
=
\tilde U_{k-1}^l+\Delta U_k^l.
\end{equation}
Here, \(\theta_v\) and \(\theta_u\) are coupling coefficients governing the adaptation dynamics.

The spike event is generated by
\begin{equation}
S_k^l = H(V_k^l-V_{\mathrm{th}}),
\end{equation}
where \(V_{\mathrm{th}}\) is the firing threshold and \(H(\cdot)\) is the Heaviside step function. 

During backpropagation, its derivative is approximated using a rectangular surrogate:
\begin{equation}
\frac{\partial S_k^l}{\partial V_k^l}
\approx
\mathbb{I}
\left(
\left|V_k^l-V_{\mathrm{th}}\right|<w
\right),
\end{equation}
where \(w\) denotes the surrogate-gradient window.

\subsection{Actor-critic architectures}
\label{app:actor_critic_architectures}

All compared methods are implemented within the standard TD3 framework \cite{fujimoto2018addressing}. The critic architecture is identical across all SNN and ANN methods and consists of two independent ANN-based Q-networks. Each Q-network takes the concatenated state-action pair \((s_t,a_t)\) as input and contains two hidden layers with 256 units and ReLU activations, followed by a scalar Q-value output. Our SpikeCredit therefore operates on the actor-side spiking dynamics and introduces temporal credit redistribution while leaving the critic architecture unchanged.

For SNN-based methods, we adopt a population-coded spiking actor based on CaRe-BN \cite{xu2026care}. Each continuous observation is first encoded into population spike trains, processed by a two-layer DN spiking MLP, and decoded into a continuous action through a population decoder. The hidden-layer membrane traces and spike events exposed by the spiking MLP serve as candidate TCCs.

For the ANN baseline, we use a conventional two-layer MLP actor with 256 hidden units per layer and ReLU activations. Its output layer uses a \(\tanh\) activation and is scaled according to the action bounds. The ANN baseline uses the same critic architecture, TD3 training procedure, and core optimization settings as the SNN baselines, but replaces the spiking actor with a non-spiking policy network.

\section{Experimental Details}
\subsection{Experiment environments}
\begin{figure*}[t!]
\centering
\includegraphics[width=0.7\textwidth]{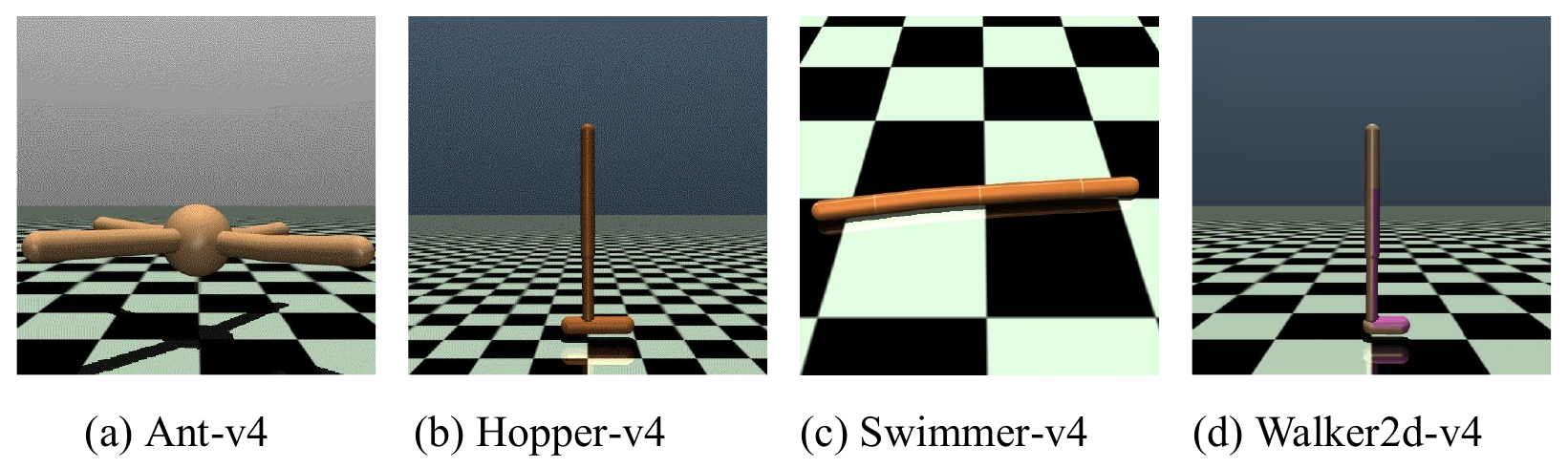} 
\caption{Rendered views of the four MuJoCo continuous-control environments used in our experiments: (a) Ant-v4, (b) Hopper-v4, (c) Swimmer-v4, and (d) Walker2d-v4.}
\label{fig:experiment_environments}
\end{figure*}

We evaluate all methods on four continuous-control tasks from the Gymnasium MuJoCo-v4 suite: Ant-v4, Hopper-v4, Swimmer-v4, and Walker2d-v4 \cite{towers2024gymnasium,todorov2012mujoco}.
As shown in Figure~\ref{fig:experiment_environments}, these environments cover locomotion tasks with different morphologies and control dynamics. Sparse rewards are constructed from the original dense rewards following the procedure described in the main text.

\subsection{Spiking neuron parameters}
\label{app:spiking_neuron_parameters}
All SNN-based methods use the same DN parameters and surrogate-gradient configuration across all tasks. As summarized in Table~\ref{tab:dn_parameters}, all neuron-dynamics parameters and the surrogate-gradient window are fixed throughout the experiments.

\subsection{Actor-critic architecture hyperparameters}
\label{app:actor_critic_architecture_hyperparameters}
All methods follow the same TD3 framework and use an identical twin-critic architecture across tasks, while differing in the actor architecture.
As summarized in Table~\ref{tab:actor_critic_architecture_hyperparameters}, all hidden-layer configurations are fixed across environments. Task-dependent observation and action dimensions affect only the corresponding input and output sizes of the actor, critic, population encoder, and population decoder.
Here, \(d_s\) and \(d_a\) denote the observation and action dimensions, respectively.

\subsection{RL and SpikeCredit hyperparameters}
\label{app:rl_spikecredit_hyperparameters}
The training, TD3, and SpikeCredit hyperparameters are summarized in Table~\ref{tab:rl_spikecredit_hyperparameters}. All tasks use the same training budget and TD3 optimization schedule. Shared SpikeCredit hyperparameters are also fixed across tasks, while \(\lambda_{\mathrm{CTT}}\) and \(\lambda_{\mathrm{sparse}}\) are task-specific. BN recalibration is applied only to SNN-based methods.

\subsection{Observation-space categorization}
For the self-motion proxy attribution analysis, we group the observation dimensions according to their physical meanings. As summarized in Table~\ref{tab:proxy_semantic_categories}, the resulting categories
describe pose, translational velocity, angular velocity, and joint-motion features, together with their abbreviations used in the main text. Table~\ref{tab:observation_semantic_grouping} provides the task-specific mapping from observation indices to these semantic categories.

The observation indices follow the default Gymnasium MuJoCo-v4 observation vectors \cite{towers2024gymnasium}. Contact-force observations are not included for Ant-v4. The field names in Table~\ref{tab:observation_semantic_grouping} are descriptive labels based on the Gymnasium observation ordering and the corresponding MuJoCo joint semantics.

For the proxy input \(z_t=[s_t,\Delta s_t,\|a_t\|_2^2]\), each dimension of \(\Delta s_t\) inherits the category of its corresponding state dimension in \(s_t\), while \(\|a_t\|_2^2\) is treated as a separate action-energy feature. These groupings are used only to aggregate and visualize feature attribution and do not affect training.

\section{Additional experiments}
\subsection{Sensitivity analysis of SpikeCredit}
\begin{figure*}[h]
\centering
\includegraphics[width=0.98\textwidth]{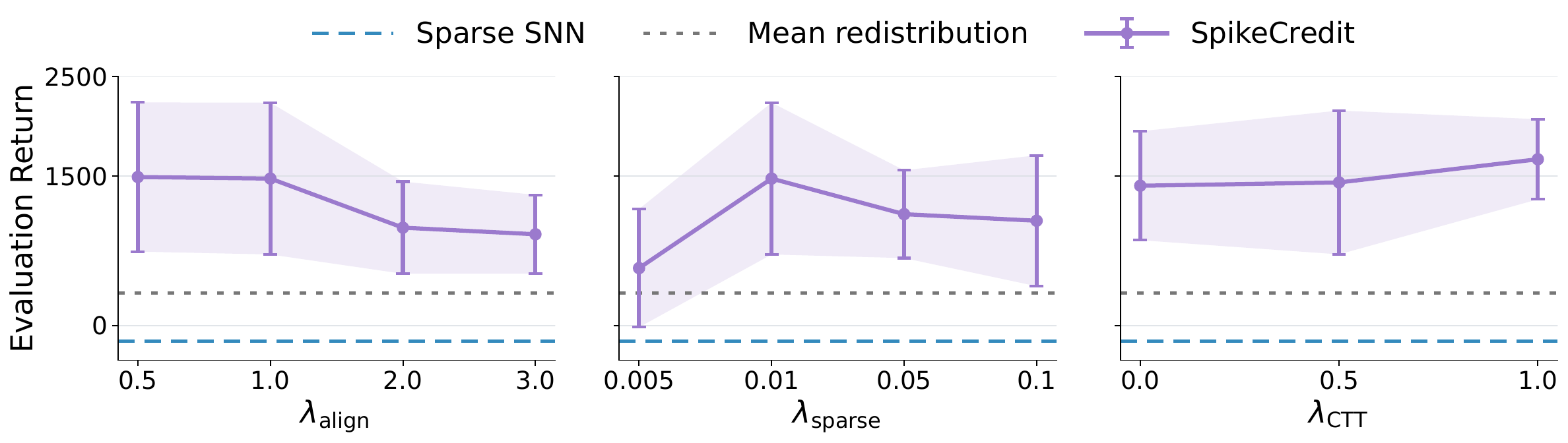}
\caption{
Hyperparameter sensitivity of SpikeCredit on Ant-v4.
We vary $\lambda_{\mathrm{align}}$, $\lambda_{\mathrm{sparse}}$, and $\lambda_{\mathrm{CTT}}$ while keeping other settings fixed, and report Last10 evaluation returns over five seeds. Dashed lines denote Sparse SNN and Mean Redistribution baselines. SpikeCredit consistently outperforms both baselines across all tested values, indicating that its gains are not tied to a narrow hyperparameter choice.
}
\label{Sensi}
\end{figure*}

Figure~\ref{Sensi} shows that SpikeCredit is robust to its main hyperparameters.
Across all tested values of $\lambda_{\mathrm{align}}$, $\lambda_{\mathrm{sparse}}$, and $\lambda_{\mathrm{CTT}}$, SpikeCredit remains above both Sparse SNN and Mean Redistribution baselines.
The best performance is obtained near moderate alignment and sparsity strengths, while overly strong alignment or overly weak sparsity reduces performance but does not collapse learning.
Increasing $\lambda_{\mathrm{CTT}}$ further improves return, supporting the role of credit-targeted trace alignment in reshaping actor dynamics into more credit-readable representations.

\clearpage

\begin{table*}[t!]
\centering
\begin{tabular}{lc}
\toprule
Parameter & Value \\
\midrule
$\alpha$ & $0.5$ \\
$V_{\mathrm{th}}$ & $0.5$ \\
$w$ & $0.5$ \\
$\theta_v$ & $-0.172$ \\
$\theta_u$ & $0.529$ \\
$\theta_r$ & $0.021$ \\
$\theta_s$ & $0.132$ \\
\bottomrule
\end{tabular}
\caption{Dynamic neuron parameters used in all SNN experiments.}
\label{tab:dn_parameters}
\end{table*}

\begin{table*}[h]
\centering
\begin{tabular}{lll}
\toprule
Component & Hyperparameter & Value \\
\midrule
TD3 critic & Number of critics & $2$ \\
TD3 critic & Input dimension & $d_s+d_a$ \\
TD3 critic & Hidden layers & $256,256$ \\
TD3 critic & Activation & ReLU \\
TD3 critic & Output dimension & $1$ \\
\midrule
SNN actor & Input / output dimension & $d_s \rightarrow d_a$ \\
SNN actor & Hidden layers & $256,256$ \\
SNN actor & Internal SNN steps $K$ & $5$ \\
SNN actor & Neuron type & DN \\
SNN encoder & Population size & $10d_s$ \\
SNN encoder & Mean range & $[-1,1]$ \\
SNN encoder & Std. & $\sqrt{0.05}$ \\
SNN encoder & Threshold & $0.999$ \\
SNN decoder & Population size & $10d_a$ \\
SNN decoder & Output activation & $\tanh$ \\
SNN layers & Normalization & enabled \\
\midrule
ANN actor & Input dimension & $d_s$ \\
ANN actor & Hidden layers & $256,256$ \\
ANN actor & Activation & ReLU \\
ANN actor & Output dimension & $d_a$ \\
ANN actor & Output activation & $\tanh$ \\
\midrule
Task dims. & Ant-v4 $(d_s,d_a)$ & $(27,8)$ \\
Task dims. & Hopper-v4 $(d_s,d_a)$ & $(11,3)$ \\
Task dims. & Swimmer-v4 $(d_s,d_a)$ & $(8,2)$ \\
Task dims. & Walker2d-v4 $(d_s,d_a)$ & $(17,6)$ \\
\bottomrule
\end{tabular}
\caption{Actor-critic architecture hyperparameters.}
\label{tab:actor_critic_architecture_hyperparameters}
\end{table*}

\begin{table*}[h]
\centering
\begin{tabular}{lll}
\toprule
Group                         & Hyperparameter                                     & Value              \\ \midrule
\multirow{6}{*}{Training}     & Random seed                                        & \(0,1,2,3,4\)      \\
                              & Batch size                                         & \(256\)            \\
                              & Replay buffer size                                 & \(10^6\)           \\
                              & Warm-up random-action steps                        & \(25{,}000\)       \\
                              & Maximum training steps                             & \(1{,}000{,}000\)  \\
                              & Evaluation interval                                & \(5{,}000\) steps  \\ \midrule
\multirow{8}{*}{TD3}          & Actor learning rate                                & \(3\times10^{-4}\) \\
                              & Critic learning rate                               & \(3\times10^{-4}\) \\
                              & Discount factor \(\gamma\)                         & \(0.99\)           \\
                              & Target update rate \(\tau\)                        & \(0.005\)          \\
                              & Exploration-noise standard deviation               & \(0.1\)            \\
                              & Target-policy-noise standard deviation             & \(0.2\)            \\
                              & Target-noise clipping range                        & \([-0.5,0.5]\)     \\
                              & Delayed policy update frequency                    & \(2\)              \\ \midrule
\multirow{2}{*}{SNN training} & BN recalibration interval                          & \(5{,}000\) steps  \\
                              & BN recalibration mini-batches                      & \(100\)            \\ \midrule
\multirow{10}{*}{SpikeCredit} & TCC scorer architecture                            & MLP                \\
                              & TCC scorer learning rate                           & \(1\times10^{-4}\) \\
                              & TCC temperature \(\tau_c\)                         & \(2.0\)            \\
                              & SMF updates per episode                            & \(1\)              \\
                              & TCC scorer gradient-norm clip                      & \(1.0\)            \\
                              & Self-motion proxy architecture                     & Linear             \\
                              & Alignment coefficient \(\lambda_{\mathrm{align}}\) & \(1.0\)            \\
                              & CTT start step                                     & \(200{,}000\)      \\
                              & Credit-target epsilon                              & \(10^{-6}\)        \\
                              & Credit-target clipping range                       & \([-5,5]\)         \\ \midrule
\multirow{2}{*}{Ant-v4}       & \(\lambda_{\mathrm{CTT}}\)                         & \(1.0\)            \\
                              & \(\lambda_{\mathrm{sparse}}\)                      & \(0.01\)           \\
\multirow{2}{*}{Hopper-v4}    & \(\lambda_{\mathrm{CTT}}\)                         & \(2.0\)            \\
                              & \(\lambda_{\mathrm{sparse}}\)                      & \(0.05\)           \\
\multirow{2}{*}{Swimmer-v4}   & \(\lambda_{\mathrm{CTT}}\)                         & \(2.0\)            \\
                              & \(\lambda_{\mathrm{sparse}}\)                      & \(0.05\)           \\
\multirow{2}{*}{Walker2d-v4}  & \(\lambda_{\mathrm{CTT}}\)                         & \(2.0\)            \\
                              & \(\lambda_{\mathrm{sparse}}\)                      & \(0.05\)           \\ \bottomrule
\end{tabular}
\caption{RL and SpikeCredit hyperparameters used in the main experiments.}
\label{tab:rl_spikecredit_hyperparameters}
\end{table*}

\begin{table*}[h]
\centering
\begin{tabular}{lll}
\toprule
Abbreviation & Category & Meaning \\
\midrule
Pose & Body / joint pose & Body height, orientation, joint angles, and foot angles \\
Fwd Vel. & Forward velocity & Translational velocity along the forward direction \\
Lat./Vert. Vel. & Lateral / vertical velocity & Sideways or vertical translational velocity \\
Body Ang. Vel. & Body angular velocity & Angular velocity of torso or body orientation \\
Joint Vel. & Joint velocity & Hip, thigh, leg, rotor, or non-foot joint velocity \\
Foot Vel. & Foot / ankle velocity & Foot or ankle joint velocity \\
\midrule
$\Delta$Pose & Change in pose & Temporal difference of Pose features \\
$\Delta$Fwd Vel. & Change in forward velocity & Temporal difference of Fwd Vel. features \\
$\Delta$Lat./Vert. & Change in lateral / vertical velocity & Temporal difference of Lat./Vert. Vel. features \\
$\Delta$Body Ang. & Change in body angular velocity & Temporal difference of Body Ang. Vel. features \\
$\Delta$Joint Vel. & Change in joint velocity & Temporal difference of Joint Vel. features \\
$\Delta$Foot Vel. & Change in foot / ankle velocity & Temporal difference of Foot Vel. features \\
Act. Eng. & Action energy & Scalar action energy $\|a_t\|_2^2$ \\
\bottomrule
\end{tabular}
\caption{Semantic proxy-feature categories and abbreviations used in the
attribution analysis.}
\label{tab:proxy_semantic_categories}
\end{table*}

\begin{table*}[h]
\centering
\begin{tabular}{llll}
\toprule
Task                         & Observation indices & Observation fields                                                                                                                      & Category        \\ \midrule
\multirow{6}{*}{Ant-v4}      & 0--12               & \begin{tabular}[c]{@{}l@{}}torso\_z; torso\_quat\_w/x/y/z; \\ hip\_1--4\_angle; ankle\_1--4\_angle\end{tabular}                         & Pose            \\
                             & 13                  & torso\_x\_velocity                                                                                                                      & Fwd Vel.        \\
                             & 14--15              & torso\_y\_velocity; torso\_z\_velocity                                                                                                  & Lat./Vert. Vel. \\
                             & 16--18              & \begin{tabular}[c]{@{}l@{}}torso\_roll\_velocity; \\ torso\_pitch\_velocity; torso\_yaw\_velocity\end{tabular}                          & Body Ang. Vel.  \\
                             & 19, 21, 23, 25      & hip\_1--4\_velocity                                                                                                                     & Joint Vel.      \\
                             & 20, 22, 24, 26      & ankle\_1--4\_velocity                                                                                                                   & Foot Vel.       \\ \midrule
\multirow{6}{*}{Hopper-v4}   & 0--4                & \begin{tabular}[c]{@{}l@{}}root\_z; root\_angle; thigh\_angle; \\ leg\_angle; foot\_angle\end{tabular}                                  & Pose            \\
                             & 5                   & root\_x\_velocity                                                                                                                       & Fwd Vel.        \\
                             & 6                   & root\_z\_velocity                                                                                                                       & Lat./Vert. Vel. \\
                             & 7                   & root\_angle\_velocity                                                                                                                   & Body Ang. Vel.  \\
                             & 8--9                & thigh\_velocity; leg\_velocity                                                                                                          & Joint Vel.      \\
                             & 10                  & foot\_velocity                                                                                                                          & Foot Vel.       \\ \midrule
\multirow{5}{*}{Swimmer-v4}  & 0--2                & body\_angle; motor1\_angle; motor2\_angle                                                                                               & Pose            \\
                             & 3                   & x\_velocity                                                                                                                             & Fwd Vel.        \\
                             & 4                   & y\_velocity                                                                                                                             & Lat./Vert. Vel. \\
                             & 5                   & body\_angle\_velocity                                                                                                                   & Body Ang. Vel.  \\
                             & 6--7                & motor1\_velocity; motor2\_velocity                                                                                                      & Joint Vel.      \\ \midrule
\multirow{6}{*}{Walker2d-v4} & 0--7                & \begin{tabular}[c]{@{}l@{}}root\_z; root\_angle; right/left thigh\_angle; \\ right/left leg\_angle; right/left foot\_angle\end{tabular} & Pose            \\
                             & 8                   & root\_x\_velocity                                                                                                                       & Fwd Vel.        \\
                             & 9                   & root\_z\_velocity                                                                                                                       & Lat./Vert. Vel. \\
                             & 10                  & root\_angle\_velocity                                                                                                                   & Body Ang. Vel.  \\
                             & 11--12, 14--15      & right/left thigh\_velocity; right/left leg\_velocity                                                                                    & Joint Vel.      \\
                             & 13, 16              & right/left foot\_velocity                                                                                                               & Foot Vel.       \\ 
\bottomrule
\end{tabular}
\caption{Task-specific mapping from Gymnasium MuJoCo-v4 observation dimensions to semantic proxy-feature categories.}
\label{tab:observation_semantic_grouping}
\end{table*}

\end{document}